\documentclass[10pt,twocolumn,letterpaper]{article}

\usepackage{wacv}
\usepackage{times}
\usepackage{epsfig}
\usepackage{graphicx}
\usepackage{amsmath}
\usepackage{amssymb}

\usepackage[ruled,vlined]{algorithm2e}
\usepackage{booktabs}
\usepackage{bbm}
\usepackage{soul} 
\usepackage{appendix}
\usepackage{multirow}
\usepackage{comment}
\usepackage{makecell}

\def\wacvPaperID{458} 

\wacvfinalcopy 

\ifwacvfinal
\def\assignedStartPage{1} 
\fi

\ifwacvfinal
\usepackage[breaklinks=true,bookmarks=false]{hyperref}
\else
\usepackage[pagebackref=true,breaklinks=true,colorlinks,bookmarks=false]{hyperref}
\fi

\ifwacvfinal
\else
\fi

\begin{document}

\title{Bottom-up Hierarchical Classification \\Using Confusion-based Logit Compression}

\author{Tong Liang, Jim Davis\\
Ohio State University\\
Columbus, Ohio 43210\\
{\tt\small \{liang.693, davis.1719\}@osu.edu}
\and
\and 
Roman Ilin\\
AFRL/RYAP\\
Wright-Patterson Air Force Base, Ohio 45433\\
{\tt\small roman.ilin.1@us.af.mil} 
}

\maketitle

\begin{abstract}
   In this work, we propose a method to efficiently compute label posteriors of a base flat classifier in the presence of few validation examples within a bottom-up hierarchical inference framework. A stand-alone validation set (not used to train the base classifier) is preferred for posterior estimation to avoid overfitting the base classifier, however a small validation set limits the number of features one can effectively use. We propose a simple, yet robust, logit vector compression approach based on generalized logits \cite{ICPR_approach} and label confusions for the task of label posterior estimation within the context of hierarchical classification. Extensive comparative experiments with other compression techniques are provided across multiple sized validation sets, and a comparison with related hierarchical classification approaches is also conducted. The proposed approach mitigates the problem of not having enough validation examples for reliable posterior estimation while maintaining strong hierarchical classification performance.
\end{abstract}

\section{Introduction}

In comparison with flat classification approaches, hierarchical classification commonly utilizes a (semantic) label hierarchy that can be manually defined or automatically derived from the semantic relationship of the set of terminal labels. This enables confident hierarchical predictions at different levels of granularity (\eg, \textit{terrier} $\rightarrow$ \textit{Dog} $\rightarrow$ \textit{Animal}) instead of enforcing flat predictions only at the terminal level with potentially low confidence. Hierarchical classification typically generalizes (bottom-up) or specifies (top-down) the initial prediction based on a given confidence threshold. Thus, estimation of reliable label posteriors is critical.

Proper posterior estimation on the output of a base classifier should employ a validation set as not to overfit the base classifier predictions on the training set \cite{ECE}. Therefore, the number of available validation examples is of particular importance. Many popular datasets either do not have a separately provided validation set or only have a small number of validation examples. Moreover, for posterior estimation techniques based on the output logits of a neural network base classifier, the associated logit vector dimensionality may be too large as compared to a smaller number of validation examples. In order to address the aforementioned issues, we propose a novel compression approach of the logits produced from a pretrained base classifier. The method harnesses relevant label confusion information from terminal predictions of the classifier on a given validation set and compresses the least confusable labels into a single derived logit value. The compressed logit vector is then used directly for label posterior estimation and subsequent hierarchical classification. Overall, the proposed method is a post-processing approach since we are modeling the confusion behavior of the base classifier instead of the inherent semantic relationship across the terminal labels. Therefore, the semantic label hierarchy is only used during inference.

\section{Related Work} \label{sect:related_work}

Hierarchical classification typically employs a label hierarchy in the form of a tree \cite{Hedging,ICPR_approach,ISVC_approach,DRTC,music_classification,face_detection_domain_knowledge} or directed acyclic graph \cite{DAG,DAG_domain_knowledge}  that explicitly injects prior knowledge of the label relationships into the model. 
The relationship between the labels can be either `IS-A' or not depending on the associated classification problem \cite{music_classification,Hedging,DAG,ICPR_approach,ISVC_approach,DRTC,face_detection_domain_knowledge,LearnHierarchy, BetterMistakes}. One could either define the label hierarchy manually with domain knowledge \cite{music_classification,DAG_domain_knowledge,face_detection_domain_knowledge} or derive the hierarchy automatically \cite{Hedging,ICPR_approach,ISVC_approach,DAG,DRTC} from a well established semantic lexical database, such as WordNet \cite{WordNet}.

There are multiple approaches that combine hierarchical classification with deep learning networks. In \cite{CNN_RNN}, a convolutional neural network (CNN) is trained for feature extraction followed by a recurrent neural network to refine the hierarchical predictions. Hierarchical losses are studied in \cite{LearnHierarchy, hierarchical_loss, BetterMistakes} for training a flat CNN classification model to study the influence of prior knowledge of the label hierarchy expressed in the training loss. Similarly, \cite{DAG} defines a conditional random field to model posteriors of the labels in a hierarchy which induced a hierarchical loss for training a common CNN model. In \cite{DRTC}, a stochastic tree sampling method is proposed to dynamically share parameters among local classifiers in a label hierarchy trained with features extracted by a CNN backend to tackle both hierarchical classification and imbalanced data classification problems. Another dynamic model based on a CNN is proposed in \cite{DSSPN} that incorporates information of a label hierarchy into the network architecture for semantic segmentation. 

In \cite{CNN_RNN,DRTC,DSSPN}, a hierarchical label is predicted in a {\em top-down} manner, where any mistakes made by a non-terminal local classifier are propagated downward to the terminal level. A combined model of both flat and top-down hierarchical classifications is proposed to tackle this issue and novelty detection in \cite{NovelDetection}. The methods proposed by \cite{Hedging,ISVC_approach,ICPR_approach} instead post-process a given flat classifier's terminal predictions by generalizing the predicted terminal labels in a {\em bottom-up} fashion through the hierarchy, where the label generalization depends on estimated label posteriors in the hierarchy.

Our proposed method is most related to the bottom-up approaches of \cite{Hedging,ISVC_approach,ICPR_approach} where no hierarchical information is used in training the base classifier. In \cite{Hedging}, a search algorithm is proposed to balance the trade-off between the specificity of the predictions and the overall hierarchical accuracy. The logit of a one-vs-rest support vector machine for each terminal label is employed for posterior estimation with Platt scaling \cite{PlattScaling}. The label posterior of a non-terminal `Super-class' is approximated via summation across all terminal descendants (assumes independence). In \cite{ISVC_approach}, a non-parametric histogram binning approach is proposed to approximate posteriors of all labels in the hierarchy, but monotonicity of the estimated posteriors along the ancestral path between a predicted terminal label and the root is not guaranteed. Unlike \cite{Hedging,ISVC_approach} that uses a single logit or softmax value of the associated label for posterior estimation, a tree-based logit vector compression approach is proposed by \cite{ICPR_approach} for label posterior estimation. Our approach employs the generalized logit technique proposed in \cite{ICPR_approach}, but does not depend on a label hierarchy and instead leverages label confusions from the base classifier.

\section{Method} \label{sect:method}

For a dataset with terminal label set \(C\), a given neural network flat classifier is denoted as \(f(x)\) with output logit vector \(L_x \in \rm I\!R^{|\it{C}|}\), i.e., $L_x=f(x)$. We refer to this flat classifier \(f\) as the `base classifier'. The proposed generalized hierarchical classifier \(h\) serves as a post-processing step after the base classifier, given by 
\begin{equation} \label{eq:g_def}
    (y,\hat{p}_y) = h(L_x,H,T)
\end{equation}
with label hierarchy {\em H} and confidence threshold \(T\:(0\leq T\leq1)\). The output is the label prediction $y$ and its posterior \(\hat{p}_y\), where $\hat{p}_y \geq T$. The prediction $y$ in Eqn.~\ref{eq:g_def} could either be a hierarchical label in $H$ or a subset of terminal labels in \(C\) depending on the choice of label generalization.

\subsection{Generalized Logit} \label{sect:generalized_logit}

A generalized logit \cite{ICPR_approach} is a derived logit for an aggregation of terminal label logits from a flat classifier. It assumes the softmax value of a Super-class label is the sum of softmax values of its terminal descendant labels in the label hierarchy. For instance, a terminal label set \(C=\{bird,dog,fish,cat,truck,car\}\) with a Super-class \textit{Vehicle} consisting of terminal labels \textit{truck} and \textit{car} has a softmax value $s_{Vehicle} = s_{truck} + s_{car} =\sum_{i\in\{\textit{truck},\textit{car}\}}s_i$. Expanding $S_{Vehicle}$ based on the logit-softmax relationship, we acquire
\begin{equation} \label{eq:s_vehicle_expand}
\begin{split}
    s_{Vehicle} &= \frac{e^{l_{truck}}}{\sum_{j\in C}e^{l_{j}}} + \frac{e^{l_{car}}}{\sum_{j\in C}e^{l_{j}}}\\ 
    & = \frac{\sum_{i\in\{\textit{truck},\textit{car}\}}e^{l_i}} {\sum_{i\in\{\textit{truck},\textit{car}\}}e^{l_i} + \sum_{j\in C/\{\textit{truck},\textit{car}\}}e^{l_j}}\\ 
    & = \frac{e^{\hat{l}_{\textit{Vehicle}}}}{ e^{\hat{l}_{\textit{Vehicle}}} + \sum_{j\in C/\textit{Vehicle}}e^{l_j}}
\end{split}
\end{equation}
where $l$ denotes the logit of a terminal label and $\hat{l}$ denotes the generalized logit of a Super-class label. The generalized logit relationship between the Super-class \textit{Vehicle} and its children is given by $e^{\hat{l}_{\textit{Vehicle}}}=\sum_{i\in\{\textit{truck},\textit{car}\}}e^{l_i}$. Therefore, the corresponding logit value of the non-terminal label \textit{Vehicle} can be derived by $\hat{l}_{\textit{Vehicle}}=ln(e^{\hat{l}_{\textit{Vehicle}}})=ln(\sum_{i\in\{\textit{truck},\textit{car}\}}e^{l_i})$. In this work, we will adopt this generalized logit formulation to acquire the logit value corresponding to a {\em subset} of terminal labels that are least confusing with the argmax predicted label.

\subsection{Compressed Logits via Label Confusion}\label{sect:tail_compression}

Unlike data-driven compression techniques that are usually designed to compress {\em large} amounts of data, we are motivated to compress the logit vector due to a {\em lack} of validation data but {\em large} logit dimensionality. We propose to construct a subset of labels (to be compressed) using the confusion matrix of the base classifier $f$ evaluated on the validation set. For a given predicted label, we aggregate the logit values of its {\em least} confused labels (not including the ground truth label) and compress them into a single generalized logit. The intuition is that logit values of the least confusing labels carry little information for posterior estimation. Furthermore, logit values preserve more residual confusion information than the exponential operation with softmax which `squashes' the probability mass of non-argmax selected classes close to 0.

The columns in the confusion matrix $M$ of $f$ correspond to the predicted labels, and the rows in $M$ are the associated ground truth labels. The aggregation procedure takes \(M_i\), the \(i^{th}\) column of \(M\) corresponding to predicted label $i$, and a given compression level \(c\) (\( 1 \leq c \leq |C|\)) as input, and outputs \(Q^{c}_i\), a list containing the ground truth label, the $(c-2)$ {\em most} confusing labels, and the single subset of the remaining {\em least} confusing labels (for label $i$). For a real application scenario, proper selection of compression level $c$ can be achieved via cross validation using the validation set (see Sect.~\ref{sect:extended_experiment}).

To acquire \(Q^{c}_i\), we first apply stable-sort (retains the order of ties) to the column vector \(M_i\) in descending order while tracking the corresponding labels (i.e., ground truth row indices in \(M_i\)). Then, we set $Q^{c}_i$ to include the first \((c-1)\) individual labels in the sorted list (ground truth label and most confusing labels for class $i$). Finally, we aggregate the remaining labels (the `tail' labels whose logits are to be compressed into a single generalized logit) into one set and append it to \(Q^{c}_i\). Employing this procedure for each column in $M$, we acquire $Q^{c}$.

For a given example $x$ and its corresponding logit vector $L_x$$=$$f(x)$, the initial label hypothesis is given by $\hat{y}$$=$$argmax(L_x)$. We then select $Q^{c}_{\hat{y}}$ from $Q^{c}$ and use the generalized logit formulation to compress the appropriate least confusing labels. For example, if \(C=\{\textit{bird},\textit{dog},\textit{fish},\textit{cat},\textit{truck},\textit{car}\}\) and \(\hat{y}=\textit{dog}\), then perhaps $Q^{c}[\hat{y}]$ $=$ $Q^{c}_{dog}$ $=$\(\{dog,cat,\{bird,fish,truck,car\}\}\) with $c$ $=$ $3$. The initial logit vector $L_x$ $=$ \([l_{\textit{bird}},l_{\textit{dog}},l_{\textit{fish}},l_{\textit{cat}},l_{\textit{truck}},l_{\textit{car}}]\) would be compressed to $L^{Q^{c}_{\textit{dog}}}_x$ $=$ \([l_{\textit{dog}},l_{\textit{cat}},\hat{l}_{\{\textit{bird},\textit{fish},\textit{truck},\textit{car}\}}]\), reducing the dimensionality from 6 to 3. The compressed logit vector $L^{Q^{c}_{\textit{dog}}}_x$ will be used in label posterior estimation and inference for the examples initially classified as \textit{dog} by base classifier.

\subsection{Label Posterior Estimation}\label{sect:posterior_estimation}

For the label posterior estimation process, we only need to estimate the posteriors of the terminal labels conditioned on the associated compressed logit vector. The posterior of any Super-class can be acquired by summing over the posteriors of its associated terminal descendants.

The posterior estimator \(\hat{p}_{j|i}\) approximates the conditional probability \(P(y=j|\hat{y}=i,L^{Q^{c}_i}_x)\), where $y=j$ denotes the ground truth of example $x$ to be label \(j\), \(\hat{y}=i\) denotes the argmax-selected base prediction for \(x\) to be label \(i\), and $L^{Q^{c}_i}_x$ is the compressed logit vector with dimension of $c$ computed from $L_x$. The posterior estimator \(\hat{p}_{j|i}\) for any $i, j \in C$ in the algorithm is modeled independently with a logistic sigmoid function given by 
\begin{equation} \label{eq:posterior_estimator}
    \hat{p}_{j|i}(L^{Q^{c}_i}_x)=\frac{1}{1+exp(-W^T_{j|i}L^{Q^{c}_i}_x + B_{j|i})}
\end{equation}
where $W_{j|i} \in \rm I\!R^{c}$ and $B_{j|i} \in \rm I\!R$ denote the parameters for the posterior estimator. To estimate \(W_{j|i}\) and \(B_{j|i}\), we employed logistic regression with the Limit-memory Broyden-Fletcher-Goldfarb-Shanno algorithm (L-BFGS-B) \cite{l_bfgs_b}.  

During posterior {\em estimation}, each terminal label \(i\) is assigned its own subset \(V_i\) of the validation set \(V\) based on the initial argmax predictions from the base classifier on $V$, i.e., \(V_i = \{x\:|\:i = argmax(L_x)\: and\:L_x \in V\}\). This argmax-selected validation subset inherently captures the confusing examples directly for the posterior estimation process. A set of posterior estimators for all terminal labels (\(\forall j \in C\)) conditioned on the same base prediction $i$ is trained with the same subset of validation data $V_i$ using the ground truth indicating whether the example actually belongs to the associated label ($j$). This is done for all combinations of terminal label posteriors ($\forall j \in C $) that are conditioned on any possible initial base prediction ($\forall i \in C$), leading to a total of $|C|\times|C|$ posterior estimators. In the unlikely case of $V_i=\emptyset$, the base classifier's terminal softmax output can be used for the label posteriors. 

For {\em inference} with test example $x$, we acquire the original logit vector from the base classifier \(L_x=f(x)\) and use its base prediction label \(\hat{y}=argmax(L_x)\) to retrieve the label aggregation result \(Q^{c}_{\hat{y}}\). Next, we compress its logit vector from $L_x$ to \(L^{Q^{c}_{\hat{y}}}_x\). Lastly, the estimated terminal label posterior value \(\hat{p}_i\) for any terminal label \(i \in C\) (previously learned using validation set $V_{\hat{y}}$) is given by
\begin{equation} \label{eq:posterior}
    \hat{p}_{i}=\frac{\hat{p}_{i|\hat{y}}(L^{Q^{c}_{\hat{y}}}_x)}{\sum_{j\in C}\hat{p}_{j|\hat{y}}(L^{Q^{c}_{\hat{y}}}_x)} 
\end{equation}
The L1-normalization is used to ensure no numerical imprecision and the probabilities across the terminal labels sum to 1. At this point we could either return the terminal label with the largest posterior (flat classification) or perform hierarchical inference.

\subsection{Bottom-Up Hierarchical Classification} \label{sect:hierarchical_classification}

The proposed hierarchical inference approach starts with checking if \( \hat{p}_{\hat{y}}\) surpasses the given confidence threshold $T$. If so, \(\hat{y}\) is returned as the prediction result along with $\hat{p}_{\hat{y}}$. Otherwise, we set our label hypothesis to the parent label of $\hat{y}$ using the semantic label hierarchy $H$. We compute the label posterior of the non-terminal label hypothesis via summation of its associated terminal descendant posteriors and check if this posterior surpasses the confidence threshold. This upward label generalization process is repeated until we find a label on the ancestral path of $\hat{y}$ with its posterior $\geq$ $T$. As the label posterior of any non-terminal label is computed from the summation of its L1-normalized terminal descendant posteriors, monotonic non-decreasing posterior inference behavior upward through the hierarchy is ensured (not guaranteed in \cite{ISVC_approach}). The overall hierarchical classification approach is provided in Algorithm~\ref{algo:inference}. 

\begin{algorithm} \label{algo:inference}
\footnotesize
\SetAlgoLined
\textbf{input}: \\
\:\:\:\:set of terminal labels $C$ \\
\:\:\:\:semantic label hierachy $H$ \\
\:\:\:\:base flat classifier $f$ \\
\:\:\:\:user-defined confidence threshold $T$\\
\:\:\:\:label posterior estimators \(\hat{p}_{j|i}\) for $i$,\:$j$ $\in$ $C$ \\
\:\:\:\:confusion aggregation $Q^{c}$ with given compression level $c$\\
\:\:\:\:testing example $x$ \\
\textbf{Output}:\\
\:\:\:\:hierarchical prediction and its estimated label posterior \((y,\hat{p}_y)\)\\
\: \\
\# get original logit and label hypothesis\\
 \(L_x = f(x)\)  \\
 \(\hat{y} = argmax(L_x)\) \\
 \# convert to compressed logits \\
 \(Q^{c}_{\hat{y}}=Q^{c}[\hat{y}]\) \\
 \(L^{Q^{c}_{\hat{y}}}_x = CompressLogits(L_x, Q^{c}_{\hat{y}})\) \\
 \# get estimated terminal label posteriors \\
 \(p_t = zeros([len(L_x),1])\) \\
 \For{\(j \in C\)}{
  \(p_t[j]=\hat{p}_{j|\hat{y}}(L^{Q^{c}_{\hat{y}}}_x)\)\\
  }
  \# L1-normalization\\
  \(p_t =p_t/sum(p_t) \) \\
 \# iteratively check labels on ancestral path from \(\hat{y}\) to root of $H$\\
 \(\hat{p}_y=0\) \\
 \(path=AncestralPath(\hat{y},H)\) \\
 \For{k in path}{
  \(td = TerminalDescendants(k,H)\) \\
  \(\hat{p}_k=sum(p_t[td])\) \\
  \If{\(\hat{p}_k \geq T\)}{
   \(y = k \) \\
   \(\hat{p}_y=\hat{p}_k\) \\
   \(return\:(y,\hat{p}_y)\) \\
  }
 }
 \caption{Bottom-up Hierarchical Classification} 
\end{algorithm}

\section{Experiments} \label{sect:res_and_dissc}

We first compared our logit compression method to alternative compression approaches across multiple validation set sizes within hierarchical classification tasks. We selected two ideal datasets that have a very large number of validation examples to fully explore and compare the approaches. CINIC-10 \cite{cinic10} has 10 terminal labels with 9K examples each for training, validation, and testing per label. Places-20 is derived from the training set of Places365-Challenge \cite{places365} that consists 20 terminal labels and has 20K training, 10K validation, and 10K testing examples per label. In our experiments, we examined the hierarchical classification performance and posterior calibration quality as the number of validation examples is {\em reduced}.

For all of the experiments on CINIC-10, we used a base flat classifier with a ResNet-20 architecture \cite{ResNetv1} trained to 82\% test accuracy. For Places-20, a WideResNet-18 \cite{wideResNet} base classifier trained to 80\% test accuracy was employed. The label hierarchy for CINIC-10 was adopted from \cite{ICPR_approach}. The label hierarchy for Places-20 is a subtree derived from the full label hierarchy of Places365-Challenge \cite{places365} and is shown in Fig.~\ref{fig:places20_tree}.

\subsection{Validation Set} \label{sect:influence_of_val_size}
We seek to evaluate our approach on different sizes of the validation set to examine its robustness against fewer examples. For CINIC-10, we randomly shuffled the original validation set \(V\) in a class-wise manner and then split it into eight validation sets (\(V^{i}, i=1,..,8\)) with reducing sizes of 9K (original size), 5K, 1K, 500, 250, 100, 50, and 25 validation examples per label. Similar shuffling and splitting was applied to the validation set of Places-20 to acquire validation sets with reducing sizes of 10K (original size), 5K, 1K, 500, 250, 100, 50, and 25 validation examples per label.

\subsection{Alternative Compression Techniques} \label{sect:res_comparison_of_compressions}

We compared our logit compression approach with the following alternative compression methods:
\begin{itemize} \small
    \item \textbf{Global PCA compression}. A global PCA transformation using all of the logit data from a given validation set is computed. Logit vectors are then compressed (projected) to the specified dimensionality (i.e., using the principal components retained). 
    
    \item \textbf{Local PCA compression}. A local PCA transformation is associated with each terminal label $i \in C$ and computed using the validation examples initially predicted as the terminal label $i$. The base prediction for any new input is used to index the corresponding local PCA transformation for logit compression to the specified dimensionality.  
    
    \item \textbf{Tree-based compression}. The tree-based logit compression method proposed by \cite{ICPR_approach} reduces the original logit vector to a smaller set containing logits of the base prediction label and (generalized) logits of labels corresponding to the upper side nodes/labels branching out from the ancestral path of the base prediction. The compressed logit vector size is dependant on both the given tree structure and the associated base prediction.
\end{itemize}

For all compression methods, we used the same label posterior estimation and hierarchical classification procedures introduced in Sect. \ref{sect:posterior_estimation}-\ref{sect:hierarchical_classification}. 

\begin{figure*}
\footnotesize
\addtolength{\tabcolsep}{-5pt}
\centering
\begin{tabular}{|c|c|c|c|c|c|c|c|c|c|c|c|c|c|c|c|c|c|c|c|}
\hline
\multicolumn{20}{|c|}{Unknown} \\
\hline
\multicolumn{7}{|c|}{Indoor} & \multicolumn{13}{c|}{Outdoor} \\
\hline
\multicolumn{3}{|c|}{Cultural} & \multicolumn{2}{c|}{Home or Hotel} & & & \multicolumn{4}{c|}{Outdoor Natural} & \multicolumn{7}{c|}{Outdoor Man Made} & & \\
\cline{1-5}\cline{8-18}
 & & & & & & &
\multicolumn{2}{c|}{\makecell{Water, \\Ice, \\Snow}} & \multicolumn{2}{c|}{\makecell{Mountains, \\ Hills, \\ Deserts, \\Sky}} & \multicolumn{3}{c|}{\makecell{Sports Fields,\\Parks,\\Leisure Spaces}} & \multicolumn{4}{c|}{\makecell{Cultural or\\Historical\\Building/Place}} & & \\ 
\cline{8-18}
\bf \makecell{aqua-\\rium} & \bf \makecell{art \\ gallery} & \bf \makecell{church \\ indoor} & \bf \makecell{bath-\\room} & \bf \makecell{bed-\\room} & \bf corridor & \bf \makecell{car\\interior} &
\bf beach & \bf coast & \bf canyon & \bf cliff & \bf  \makecell{amuse-\\ment\\park} & \bf \makecell{athletic\\field} & \bf \makecell{ baseball\\field} & \bf arch & \bf castle & \bf \makecell{cemet-\\ery} & \bf \makecell{church\\outdoor} & \bf bridge & \bf \makecell{building\\facade} \\ 
\hline
\end{tabular}
\caption{Label hierarchy for Places-20.}
\label{fig:places20_tree}
\end{figure*}

\subsection{Evaluation Metrics} \label{sect:eval_metrics}

We employed the bottom-up hierarchical evaluation metrics from \cite{ISVC_approach,ICPR_approach} that focus on the two resulting subsets from test examples that were either correctly (\(S_c\)) or incorrectly (\(S_{ic}\)) classified by the base classifier $f$ and also a measure that correlates with their label depth in the tree:

\begin{itemize} \small
  \item \textbf{C-persist}, \textbf{C-corrupt}, \textbf{C-withdrawn} are the fractions of the correct base predictions ($S_c$) that are unchanged as terminal labels, corrupted to incorrect predictions, or assigned to the whole set of terminal labels $C$ (i.e., the root \textit{Unknown}), respectively. Unlike \cite{Hedging}, our hierarchical classification approach does not corrupt any base predictions (C-corrupt = 0). \textbf{C-soft} is the remaining fraction in $S_c$ being generalized to a non-root Super-class that has the ground truth label as a descendant.  
  
  \item \textbf{IC-persist} and \textbf{IC-withdrawn} are similar to C-persist and C-withdrawn but for the incorrect base predictions ($S_{ic}$). \textbf{IC-reform} is the fraction of base predictions in $S_{ic}$ that are generalized to a {\em correct} non-root label. \textbf{IC-remain} is the remaining fraction of examples in $S_{ic}$ being generalized to a non-root Super-class that does {\em not} have the ground truth as a descendant.
  
  \item \textbf{Avg-sIG} is the average scaled Information Gain, adapted from \cite{Hedging}'s metric of Information Gain (IG). Scaled Information Gain (sIG) is IG normalized with the maximum possible IG for a given terminal label set, and Avg-sIG is the average sIG across all test examples and correlates with the average label depth. Avg-sIG is given by
  \begin{equation} \label{eq:sIG}
      \text{Avg-sIG}=\frac{1}{|S|} \sum_{x\in S} \frac{log_2 |C|-log_2|y|}{log_2|C|}\mathbbm{1}_{\{t_x \in y\}}
  \end{equation}
  where $y$ is the final prediction of input $x$,  $|y|$ is the number of terminal descendants of $y$, $t_x$ is the ground truth of test example $x$, $S=S_c\cup S_{ic}$, and $\mathbbm{1}_{\{.\}}$ is the indicator function. An incorrect prediction has Information Gain of 0, as defined in \cite{Hedging}.
\end{itemize}

We adopted the \textbf{Technique of Order Preference Similarity to the Ideal Solution (TOPSIS)} \cite{TOPSIS} as an overall evaluation metric that unifies the 9 individual hierarchical metrics for the comparison across alternative approaches. TOPSIS is a commonly used metric for multiple criteria decision analysis. It treats the multi-criteria evaluation results (\eg, the 9 hierarchical metrics listed above) of an approach as vector \(a\) and computes the L2-distance to the best alternative ($b$) and worst alternative ($w$) vectors. The final TOPSIS score for $a$ ranges from zero to one (larger is better) and is given by
\begin{equation} \label{eq:topsis}
  s_{a}=\frac{d_{aw}}{d_{aw}+d_{ab}}
\end{equation}
where $d_{ab}$ and $d_{aw}$ are the L2-distances from $a$ to the best ($b$) and worst ($w$) alternatives, respectively. 

For the best case scenario ($b$), C-persist = IC-reform = Avg-sIG = 1, and the remaining criteria/metrics are 0. For the worst case scenario ($w$), C-corrupt = IC-persist = 1, and the remaining criteria are 0. We note that the metric/criterion values are typically rescaled in the range between 0 and 1 and can be weighted before computing the L2-distances. Our hierarchical metrics are naturally defined within the same range and are treated equally.

Lastly, we evaluated the calibration quality of the estimated label posteriors with \textbf{Expected Calibration Error (ECE)} \cite{ECE}, as the inference approach is highly dependent on the label posteriors. ECE is a scalar statistical summary of the quality of the estimated label posteriors. This is evaluated through histogram binning of the posterior values and comparing the average of the estimated label posteriors in each bin to the associated precision of the hierarchical predictions in the bin. Since our hierarchical predictions are acquired under a given confidence threshold, we apply the histogram binning for ECE on the probability range between the confidence threshold $T$ and 1 (with 10 bins), and it is only applied to predictions that are not withdrawn/generalized to the root of the hierarchy (which always has posterior of 1). Lower values of this metric are preferred.

\begin{figure*}[t]
    \centering
    \setlength{\tabcolsep}{0.0pt}
    \begin{tabular}{cccc}
    Proposed & Global PCA & Local PCA & Tree-Based \cite{ICPR_approach} \\
    \toprule
    \includegraphics[height=1.3in]{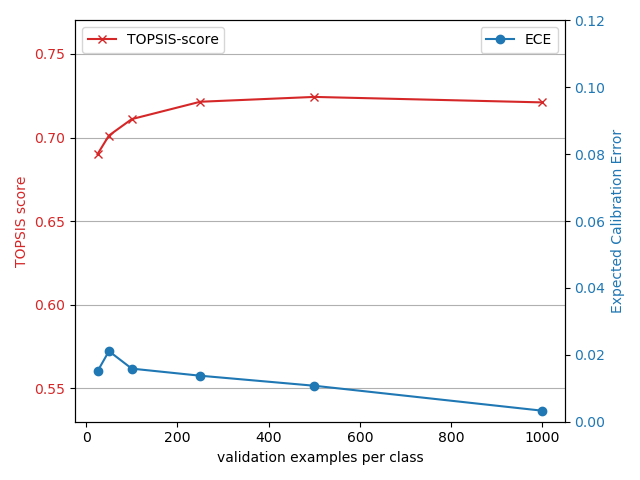} &
    \includegraphics[height=1.3in]{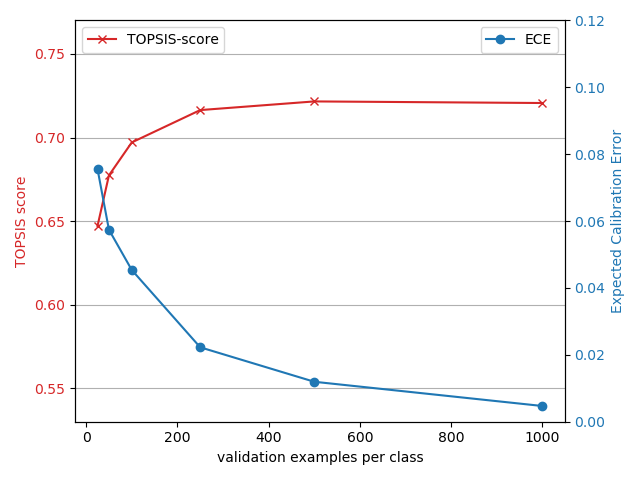} & 
    \includegraphics[height=1.3in]{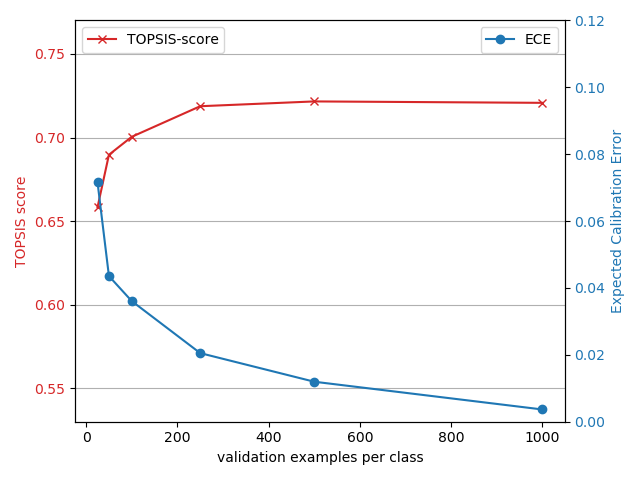} & 
    \includegraphics[height=1.3in]{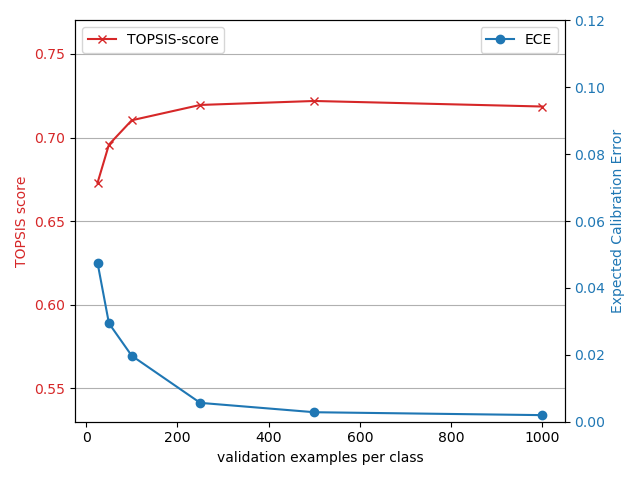}  \\
    \multicolumn{4}{c}{(a) CINIC-10} \\
    
    \includegraphics[height=1.3in]{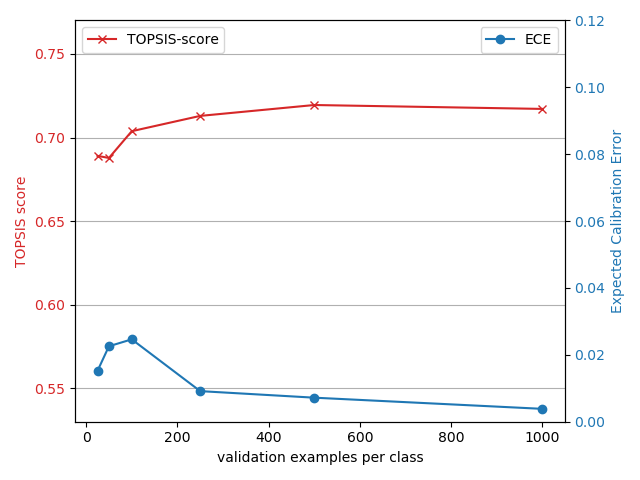} &
    \includegraphics[height=1.3in]{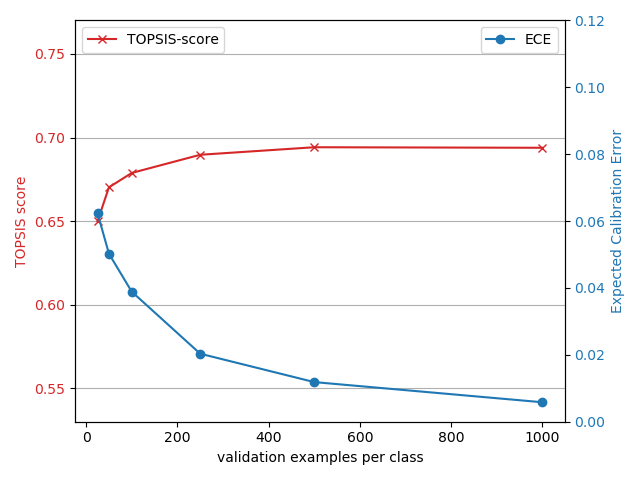} & 
    \includegraphics[height=1.3in]{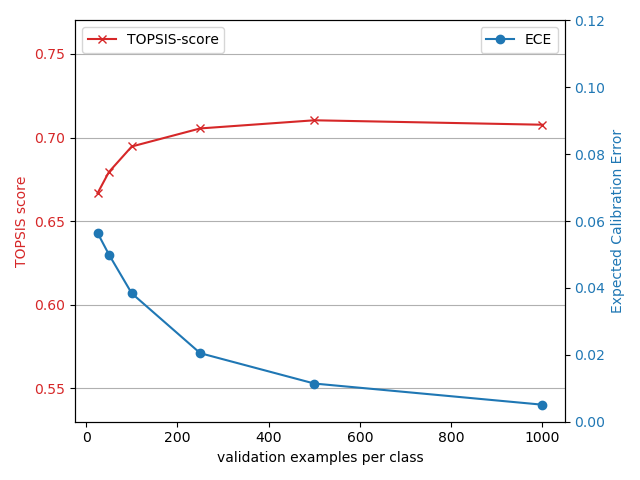} & 
    \includegraphics[height=1.3in]{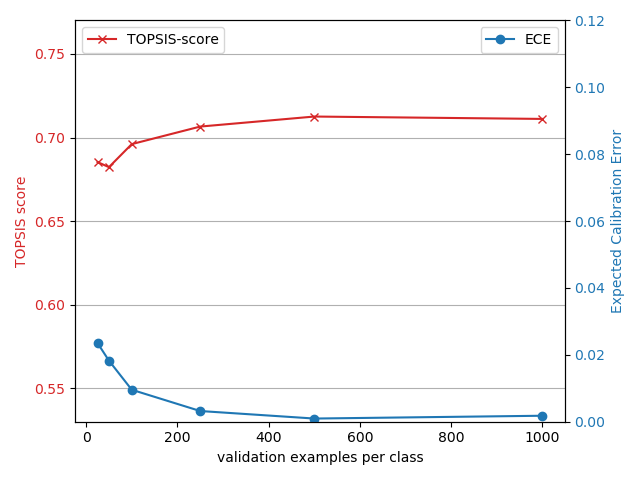}  \\
    \multicolumn{4}{c}{(b) Places-20} \\
    \bottomrule
    \end{tabular}
    \caption{Comparison of compression techniques on (a) CINIC-10 and (b) Places-20 with normal semantic tree.}
    \label{fig:Compression_tech_comparison_normal_tree}
\end{figure*}

\begin{figure*}[t]
    \centering
    \setlength{\tabcolsep}{0.4pt}
    \begin{tabular}{c c c c}
    Proposed & Global PCA & Local PCA & Tree-Based \cite{ICPR_approach} \\
    \toprule

    \includegraphics[height=1.3in]{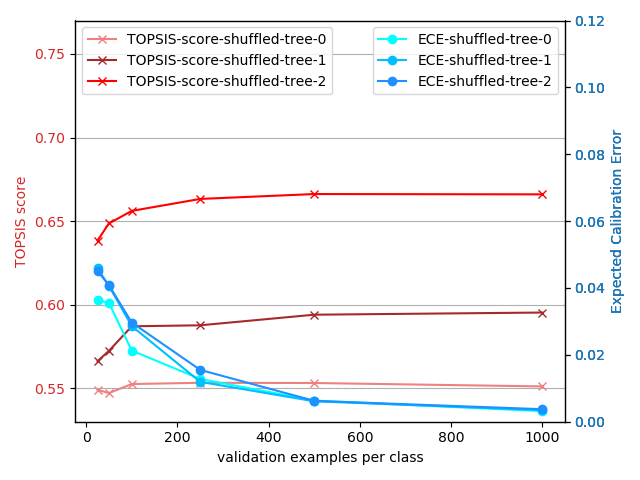} &
    \includegraphics[height=1.3in]{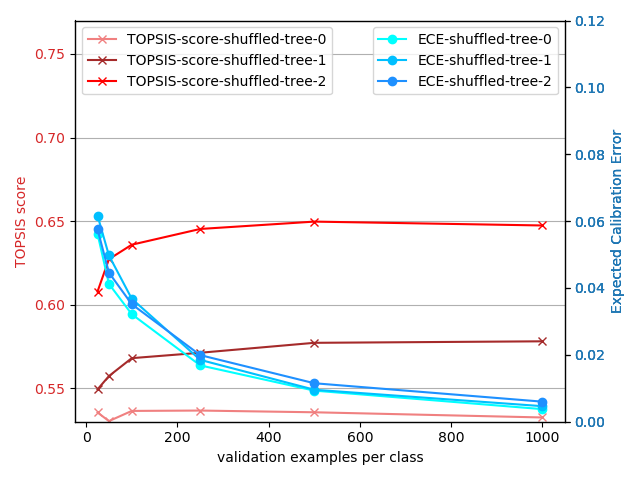} & 
    \includegraphics[height=1.3in]{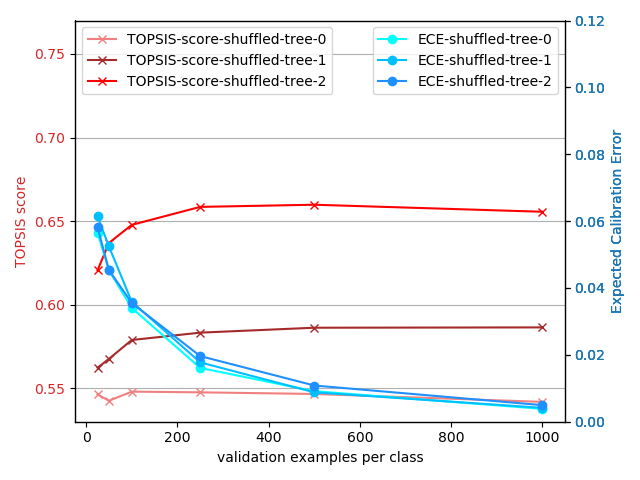} & 
    \includegraphics[height=1.3in]{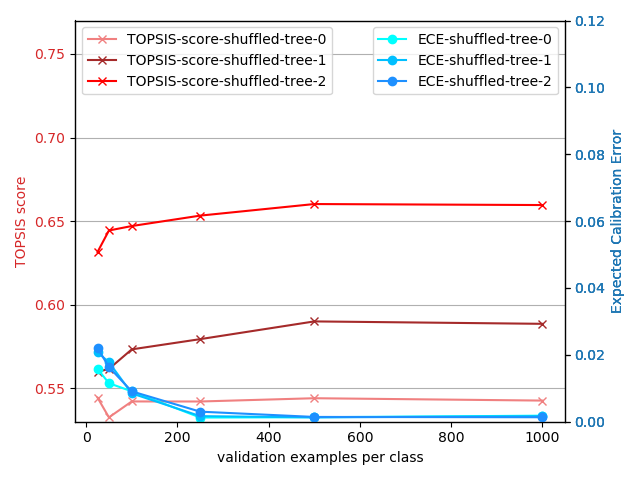}  \\
    \bottomrule
    \end{tabular}
    \caption{Comparison of compression techniques on Places-20 with {\em shuffled} trees.}
    \label{fig:Compression_tech_comparison_shuffled_tree}
\end{figure*}

\begin{figure*}[t]
    \centering
    \setlength{\tabcolsep}{0.4pt}
    \begin{tabular}{c c c}
    Proposed & Global PCA & Local PCA \\
    \toprule
    
    \includegraphics[height=1.3in]{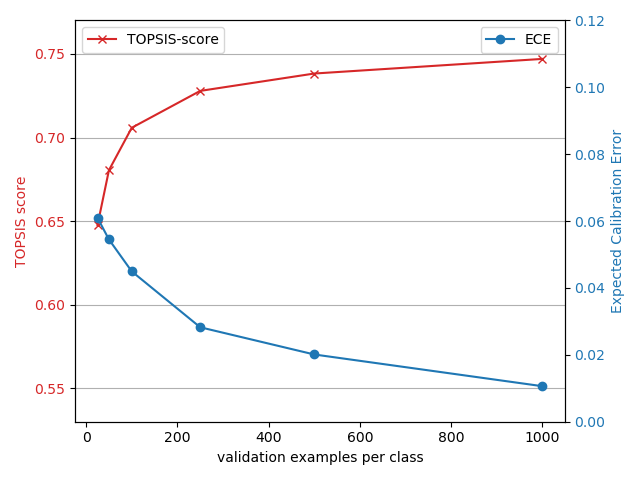} &
    \includegraphics[height=1.3in]{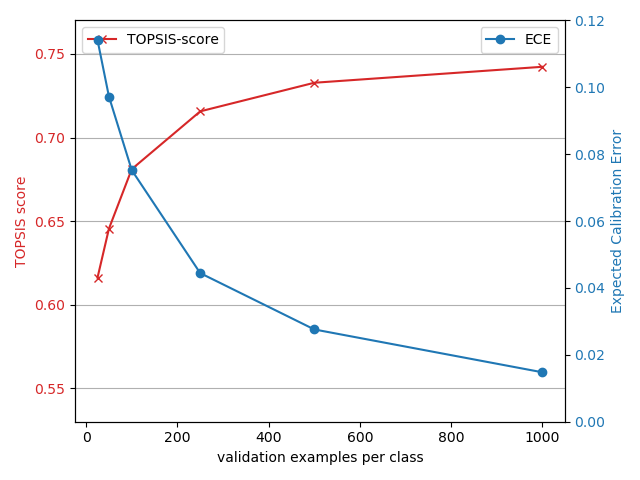} & 
    \includegraphics[height=1.3in]{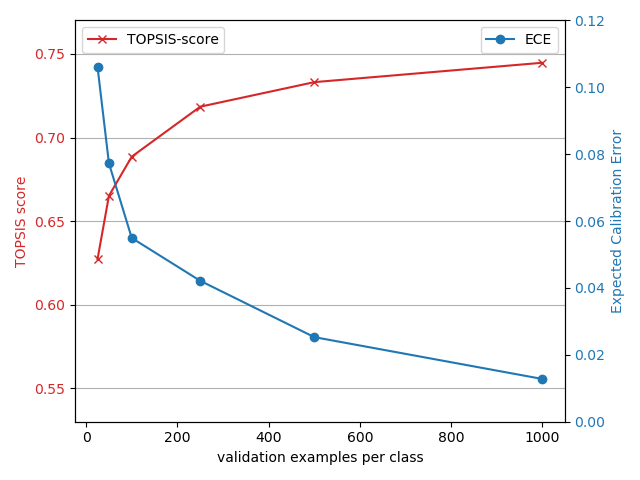} \\
    \multicolumn{3}{c}{(a) CINIC-10} \\
    
    \includegraphics[height=1.3in]{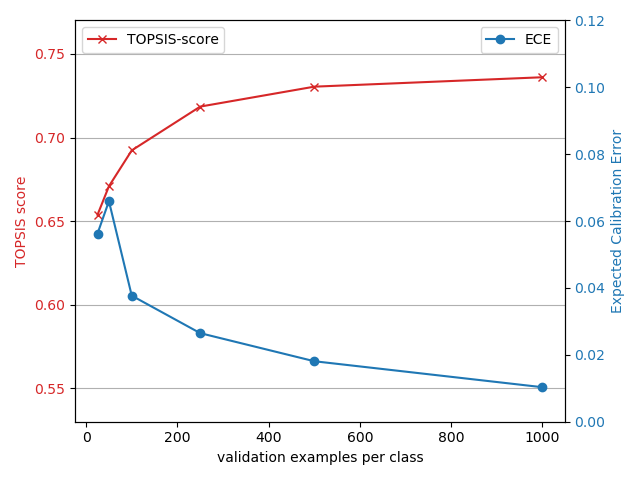} &
    \includegraphics[height=1.3in]{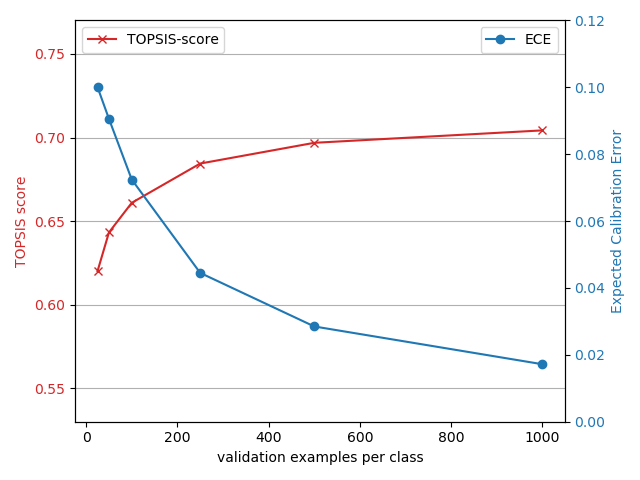} & 
    \includegraphics[height=1.3in]{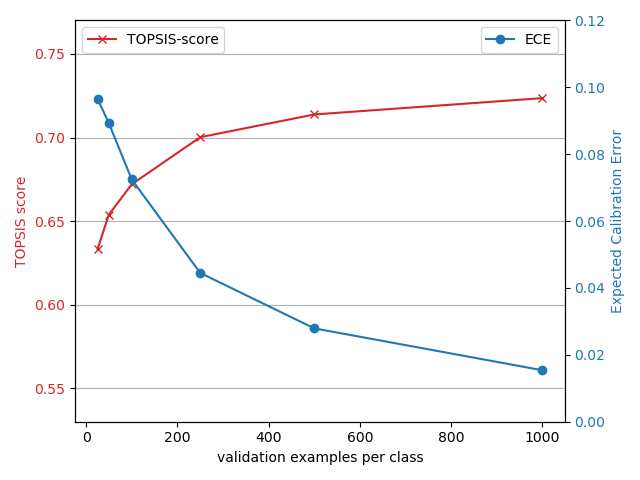}  \\
    \multicolumn{3}{c}{(b) Places-20} \\
    
    \bottomrule
    \end{tabular}
    \caption{Comparison of compression techniques on (a) CINIC-10 and (b) Places-20 with {\em no} tree label generalization.}
    \label{fig:Compression_tech_comparison_no_tree}
\end{figure*}

\subsection{Results}

We compared the results on CINIC-10 and Places-20 under various compression techniques with decreasing sizes of the validation set and report the TOPSIS and ECE scores of each method using the confidence threshold $T=0.9$. We show the TOPSIS and ECE scores of each method that corresponded to the compression level giving the best TOPSIS results (ranging from 1 to 10). To automatically select the compression level in an application scenario, cross validation could be used (to be demonstrated in Sect.~\ref{sect:extended_experiment}). We note that the tree-based compression method does not support different choices of compression level, as it is specified by the tree itself.

The comparison results for experiments on CINIC-10 and Places-20 are shown in Fig.~\ref{fig:Compression_tech_comparison_normal_tree}. The left y-axis of each subplot denotes the TOPSIS score (red), and the right y-axis denotes the associated ECE score (blue). The x-axis of each subplot denotes the size of the validation set used to acquire the associated TOPSIS and ECE scores. Axes of all subplots are given in the same range to make it easier for visual comparison. We only show the results derived from the smaller validation sets ($V^{3}$ -- $V^{8}$) since all methods stabilized with any further increase of the validation data.

Notably, the results for TOPSIS and ECE derived from our proposed compression method and the tree-based compression approach are better than both PCA-based approaches when the validation set is small (i.e., 25, 50, 100 examples per label). The results of our method is better than the tree-based approach on CINIC-10 and on par with it for Places-20. This suggests that our compression approach derived from the confusion matrix of the base classifier is comparable to the tree-based compression approach that has injected knowledge of the semantic label relationships. Both methods employ extra knowledge (classifier confusions or label semantics) and have better performance than pure data driven PCA-based approaches when the validation set is small. Our method reached the highest TOPSIS score before saturation among the methods. Similar saturation is also observed for the ECE scores, where the tree-based compression method first reached saturation, with the lowest ECE score, followed closely by our compression approach. However, our method had the best ECE score for the smallest validation set in CINIC-10.

Since all four methods in the comparison rely on a given semantic hierarchy for their hierarchical classification, we next conducted experiments with the same settings but randomly corrupted the semantic hierarchies to study the influence of the tree itself. We applied a random shuffling of the terminal label positions only within each hierarchy while keeping the tree structure fixed. The results for 3 random terminal shufflings of the hierarchy for Places-20 are shown in Fig.~\ref{fig:Compression_tech_comparison_shuffled_tree}. Comparing the results from the original label tree with the shuffled trees, the semantic information of the label relationships provided by the hierarchy does indeed play an important role in hierarchical classification, as degraded performance for all methods is present if an improper tree is used. Notably, the ECE scores for our approach and the tree-based method are still better than the PCA-based approaches for smaller validation sets. Similar results were seen with shuffled trees on CINIC-10.  

A similar comparative experiment among these methods {\em without} the use of a hierarchy was also conducted. The tree-based compression approach is not applicable here due to its requirement of a tree, so it is excluded from this comparison. We employed a basic label generalization method that takes the estimated label posteriors of all terminal labels and sorts them in descending order, then sums the sorted posteriors until reaching the given confidence threshold $T$. The {\em set} of aggregated terminal labels is then output as a (generalized) hierarchical prediction (i.e., a subset of terminal labels in $C$). Our evaluation metrics are still applicable to these set-based predictions. The results on CINIC-10 and Places-20 are shown in Fig.~\ref{fig:Compression_tech_comparison_no_tree}. Comparing results to Fig.~\ref{fig:Compression_tech_comparison_normal_tree}, the TOPSIS and ECE scores for each approach without using a tree are worse for smaller validation sets, though the proposed approach is again better than the PCA methods. Additionally, the TOPSIS scores do not saturate as much as before (when employing a tree for inference). 

\begin{table}[t]
    \centering
    \small\addtolength{\tabcolsep}{-2pt}
    \begin{tabular}{||c||c||c|c||}
        \toprule
         configuration                & original    & worst     & best \\
         \midrule
         25 val. examples per class   & .633 (10)  & .630 (1)  & .690 (\textbf{2}) \\ 
         50 val. examples per class   & .650 (10)  & .644 (1)  & .701 (\textbf{3}) \\ 
         100 val. examples per class  & .660 (10)  & .656 (1)  & .711 (\textbf{3}) \\
         250 val. examples per class  & .673 (10)  & .670 (1)  & .721 (\textbf{6}) \\
         \bottomrule
    \end{tabular}
    \caption{TOPSIS scores and the number of compressed logits used for hierarchical classification at 90\% confidence threshold. The number in parenthesis is the associated compression level used.}
    \label{tab:ablation}
\end{table}

We conducted an ablation study on the influence of the proposed confusion-based compression approach by comparing the associated hierarchical classification performance with and without the proposed compression method on CINIC-10 when the validation set is small (i.e., 25, 50, 100, and 250 examples per label). For the proposed approach, we show its best and worst TOPSIS scores across compression levels ranging from 1 to 9 in Table~\ref{tab:ablation}. In comparison, we also show the TOPSIS score of hierarchical classification using the original logit features with no compression. The results clearly show that the proposed compression approach improves the hierarchical classification performance significantly in comparison with using the original logits without compression. The results also indicate that a smaller validation set will likely require a smaller compression level to reach better performance.

\subsubsection{Discussion}

The use of label hierarchy appears to have a regularization effect on the inference procedure. The semantic relationship of labels is therefore helpful during inference when the given validation set is relatively small, but tree-based inference becomes more of a constraint and limits the performance (as compared with label generalization without a tree) when a sufficient amount of validation examples is available. Therefore, the proposed overall hierarchical classification framework with compressed logits is most suitable when relatively fewer validation examples are given or when a label hierarchy is unavailable.

\subsection{Scalability} \label{sect:extended_experiment}

We next compared our proposed framework against the related bottom-up hierarchical classification methods of \cite{Hedging}, \cite{ISVC_approach}, and \cite{ICPR_approach} (described in Sect.~\ref{sect:related_work}) under the same confidence threshold $T=0.9$. The experiments were conducted on Fashion-MNIST \cite{fashionmnist} (10 terminal labels),  CIFAR-100 \cite{CIFAR} (100 terminal labels), and ImageNet-Animal \cite{ISVC_approach} (398 terminal labels) with increasing scale of the dataset and the related label hierarchy (provided in \cite{ICPR_approach}). We applied a similar random partitioning on the original test sets of Fashion-MNIST and CIFAR-100 to obtain the associated validation and test sets. The validation and test sets for ImageNet-Animal were acquired by class-wise random 1:1 partition of the original validation set, since no test set is given. The validation sets of all three datasets were balanced with 500, 50, and 25 validation examples per terminal label for Fashion-MNIST, CIFAR-100, and ImageNet-Animal, respectively. A model based on VGGNet \cite{VGGNet} was used for Fashion-MNIST, a ResNeXt model \cite{ResNext} was trained and used for CIFAR-100, and a pretrained ResNet-152 model \cite{ResNetv1} was employed as the base classifier for ImageNet-Animal.

For \cite{Hedging}, the single logit feature of each label from the base (CNN) classifier was used to produce the hierarchical predictions instead of logits of SVMs used in the original work. For \cite{ISVC_approach}, we employed the setting of 10 histogram bins per label. To automatically determine the proper compression level for our method on each dataset, we applied class-wise Leave-One-Out Cross Validation (LOOCV) on each validation set. The search range was from 1 to the minimum of $|C|$ and the number of validation examples per label. The compression level that led to the highest TOPSIS score for each dataset was 5, 9, and 12 for Fashion-MNIST, CIFAR-100, and ImageNet-Animal, respectively. These compression levels were employed by our approach to produce the final hierarchical classification results. 

The TOPSIS scores corresponding to each method are shown in Table \ref{tab:extended_comparison_3_datasets}, where `Base' refers to the flat classification results of the associated base classifiers.~The performance of our proposed approach reached the highest TOPSIS scores on all 3 datasets. Across the datasets, as the number of validation examples {\em reduced} (Fashion-MNIST $>$ CIFAR-100 $>$ ImageNet-Animal), the number of terminal labels {\em increased}. With this inverse relationship, the condition for posterior estimation is worsening. However, our proposed approach demonstrated robustness in each case.

The non-parametric histogram binning approach \cite{ISVC_approach} demonstrated good performance closest to our proposed method across the datasets, though monotonicity of label posteriors along the hierarchy is not guaranteed in this approach. The lower performance of \cite{Hedging} was due to its consistently high IC-persist, indicating that more of the original misclassifications made by the base classifier were not corrected in their hierarchical process. The lower TOPSIS scores of the tree-based compression approach \cite{ICPR_approach} on CIFAR-100 and ImageNet-Animal were due to a higher overall label generalization during inference (i.e., high C-soft and corresponding low Avg-sIG). We found that using argmax-selected validation data (as employed in the proposed approach) for \cite{ICPR_approach} would significantly improve its performance, though \cite{ICPR_approach} still requires the use of a specified hierarchy for both compression and inference, unlike the proposed approach.

\begin{table}[t]
    \centering
    \small\addtolength{\tabcolsep}{-2pt}
    \begin{tabular}{||c||c|c|c|c|c||}
        \toprule
         Dataset         & Base   & \cite{Hedging} & \cite{ISVC_approach} & \cite{ICPR_approach} & Ours\\
         \midrule
         Fashion-MNIST (10)  & .544  & .615 & .844  & .821 &\textbf{.852} \\ 
         CIFAR-100 (100)      & .534 & .690 & .735  & .672 &\textbf{.744} \\ 
         ImageNet-Animal (398) & .537  & .697 & .715 & .569   &\textbf{.720}\\
         \bottomrule
    \end{tabular}
    \caption{TOPSIS scores of hierarchical classification results at 90\% confidence threshold. The number in the parenthesis indicates the number of terminal labels.}
    \label{tab:extended_comparison_3_datasets}
\end{table}

\section{Conclusion} \label{sect:conclusion}

In the context of hierarchical classification when posterior estimation from limited validation examples is encountered, we proposed a robust and effective logit vector compression approach to preserve as much useful information as possible in the original logit vector from the perspective of label confusions. By examining the confusion matrix of a given flat classifier, we constructed a subset of the terminal labels that are least confusing with respect to each terminal ground truth label and employed the generalized logit formulation \cite{ICPR_approach} to compress the associated logits. We compared the proposed approach to data-driven compression techniques and a tree-based compression technique on the CINIC-10 and Places-20 datasets under a) different sizes of the validation set, b) randomization in the label hierarchy, and c) absence of a label hierarchy. The strong performance of our proposed approach was demonstrated across the evaluations and maintained robust performance in extended experiments comparing with related hierarchical classification approaches on datasets Fashion-MNIST, CIFAR-100, and ImageNet-Animal scaling from small to large.

\section*{Acknowledgement}
This research was supported by the U.S. Air Force Research Laboratory under Contract \#GRT00054740.

{\small
\bibliographystyle{ieee_fullname}
\bibliography{egbib}
}

\end{document}